\documentclass[10pt,twocolumn,letterpaper]{article}

\usepackage[pagenumbers]{cvpr} 
\usepackage{url}
\usepackage{times}
\usepackage{epsfig}
\usepackage{graphicx}
\usepackage{amsmath}
\usepackage{amssymb}
\usepackage{multirow}
\usepackage{booktabs}
\usepackage{setspace}
\usepackage{color}
\usepackage{algorithm}
\usepackage{algpseudocode}
\usepackage{algorithmicx}
\usepackage{amsmath}
\usepackage{threeparttable}
\renewcommand{\algorithmicrequire}{\textbf{Input:}}  
\renewcommand{\algorithmicensure}{\textbf{Output:}} 

\usepackage[pagebackref=true,breaklinks=true,colorlinks,bookmarks=false]{hyperref}
\usepackage[table,xcdraw]{xcolor}

\def\cvprPaperID{5582}
\def\confName{CVPR}
\def\confYear{2022}
\begin{document}
\title{Robust Representation via Dynamic Feature Aggregation}
\author{Haozhe Liu$^{1,2}$, Haoqin Ji$^{1,2}$, Yuexiang Li$^{2*}$, Nanjun He$^2$, Haoqian Wu$^1$, Feng Liu$^{1}$, \\
Linlin Shen$^1$, Yefeng Zheng$^2$ \\
$^1$ Computer Vision Insitute, Shenzhen University; \\
$^2$Jarvis Lab, Tencent. Shenzhen, China; \\
}
\maketitle

\begin{abstract}
    Deep convolutional neural network (CNN) based models are vulnerable to the adversarial attacks. One of the possible reasons is that the embedding space of CNN based model is sparse, resulting in a large space for the generation of adversarial samples. In this study, we propose a method, denoted as Dynamic Feature Aggregation, to compress the embedding space with a novel regularization. Particularly, the convex combination between two samples are regarded as the pivot for aggregation. In the embedding space, the selected samples are guided to be similar to the representation of the pivot. On the other side, to mitigate the trivial solution of such regularization, the last fully-connected layer of the model is replaced by an orthogonal classifier, in which the embedding codes for different classes are processed orthogonally and separately. With the regularization and orthogonal classifier, a more compact embedding space can be obtained, which accordingly improves the model robustness against adversarial attacks. An averaging accuracy of 56.91\% is achieved by our method on CIFAR-10 against various attack methods, which significantly surpasses a solid baseline (Mixup) by a margin of 37.31\%. More surprisingly, empirical results show that, the proposed method can also achieve the state-of-the-art performance for out-of-distribution (OOD) detection, due to the learned compact feature space. An F1 score of 0.937 is achieved by the proposed method, when adopting CIFAR-10 as in-distribution (ID) dataset and LSUN as OOD dataset. Code is available at \url{https://github.com/HaozheLiu-ST/DynamicFeatureAggregation}.
\end{abstract}

\section{Introduction}
The recent advances in convolutional neural networks (CNNs) have yielded significant improvements to various computer vision tasks, such as image classification \cite{Liu_2021_ICCV,he2016deep,szegedy2017inception} and image generation \cite{goodfellow2014generative,liu2021manifold,isola2017image}. However, recent studies \cite{shafahi2019adversarial,hendrycks2019benchmarking} revealed that CNN based models are vulnerable to the adversarial attacks and out-of-distribution samples, which seriously limits their applications in security-critical scenarios, \emph{e.g.,} self-driving \cite{tu2021exploring}, medical imaging\cite{ma2021understanding} and biometrics \cite{liu2021taming,liu2021one}.

\begin{figure}[!tb]
    \centering
    \includegraphics[width=.48\textwidth]{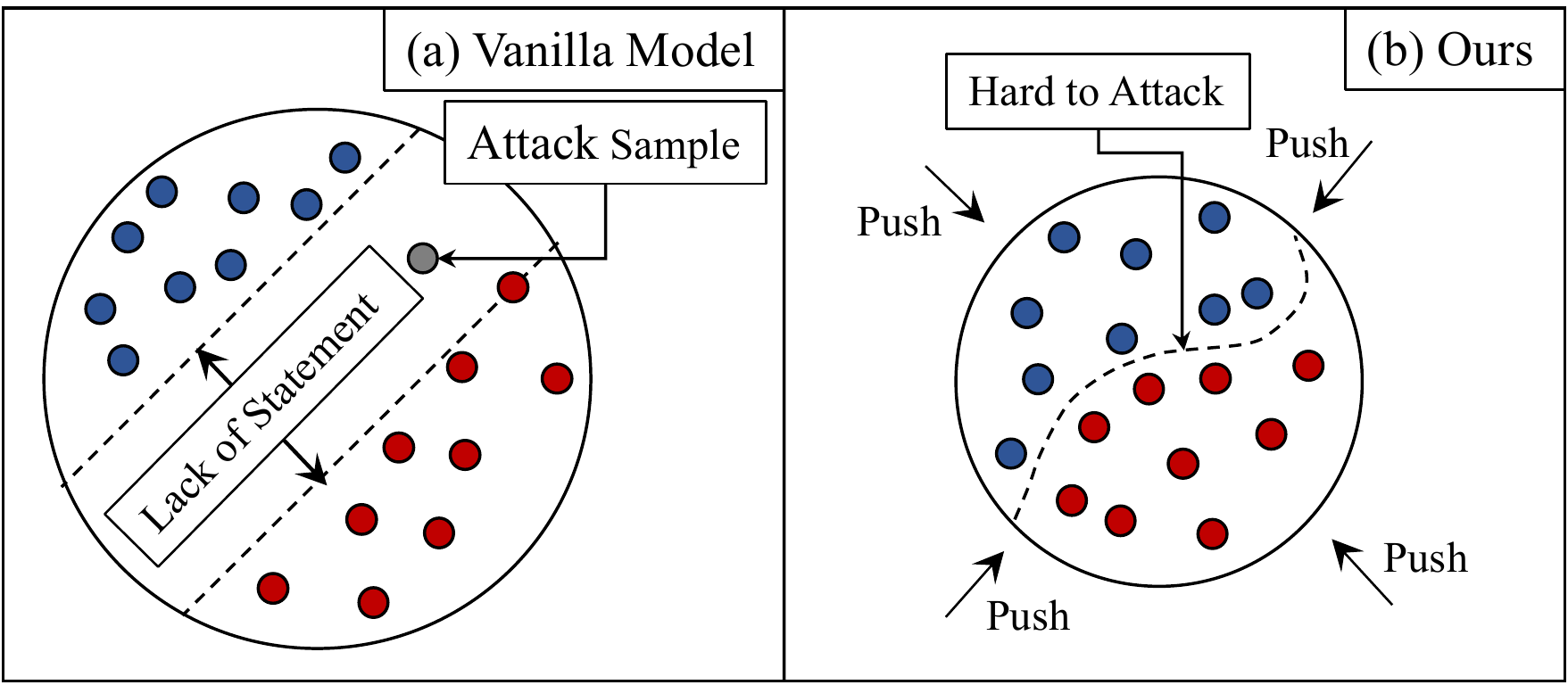}
    \caption{Our solution to improve the robustness of vanilla (traditional) CNN based model. Compared to the vanilla model (a), the proposed solution (b) drives the model to aggregate the extracted features into a compact space. Since each point in embedding space has clear statement, adversarial attacks cannot find a proper point to attack.}
    \label{fig:motivation}
\end{figure}

Such an open challenge raises increasing concerns from the community, and then numerous solutions have been proposed. One of the typical methods is adversarial training \cite{xie2019feature,shafahi2019adversarial,zheng2020efficient}, which uses the attack samples for network training to improve the model robustness.
However, adversarial training has some unexpected issues, \emph{i.e.,} high computational cost and over-fitting. Since adversarial images require repeated gradient computations, the time consumption of each training iteration is often over 
three times longer than vanilla training \cite{shafahi2019adversarial}.
Furthermore, adopting adversarial images for training inevitably introduces a bias on non-adversarial images. In particular, the model pays more attentions to the adversarial attacks while neglects the features from clean images. Therefore, adversarial training based models cannot generalize well to clean images. Recently, Rice \emph{et al.} \cite{rice2020overfitting} proved that, after a specific point, models with adversarial training can continue to decrease the training loss, while increasing the test loss. Such a phenomenon indicates that the features from different adversarial samples are diverse, which is difficult for models to learn the common patterns. To this end, improving model robustness without adversarial samples becomes a more promising research line for the challenge.

One of the representative non-adversarial methods is Mixup \cite{zhang2017mixup}, which can simultaneously improve generalization and robustness of models by generating linear combinations of pair-wise examples for model training.
Zhang \emph{et al.} \cite{zhang2020does} theoretically proved that Mixup refers to an upper bound of the adversarial training, which explained the reason behind the improvement of model robustness yielded by Mixup. However, the upper bound is not a strong constraint for CNN based model, \emph{i.e.,} Mixup generally obtains the competitive performance against single-step attacks (\emph{e.g.,} FGSM \cite{goodfellow2014explaining}), but cannot defense more challenging multi-step attacks (\emph{e.g.,} PGD \cite{athalye2018obfuscated,madry2017towards}) effectively.

To tackle the problem, we propose a new solution, called Dynamic Feature Aggregation, which can improve the robustness against strong attacks (\emph{e.g.,} PGD \cite{athalye2018obfuscated,madry2017towards} and CW \cite{carlini2017towards}), without using the adversarial samples or sacrificing the generalization capacity. As shown in Fig. \ref{fig:motivation}, 
the main idea behind our method is to drive
CNN based models to learn a compact feature space for classification. Since the data distributions of different categories are dense around the decision boundary, no space is left to generate adversarial attacks, which results in the improvement of the model robustness. Concretely, the linear interpolation with two samples is regarded as a pivot, and then adopted as the center for aggregation in the embedding space. By aggregating given samples towards the center, a compact embedding space can be thereby obtained. In order to alleviate trivial solutions caused by over-aggregating, we replace the last fully-connected layer of CNN with an orthogonal classifier to further ensure the feature discrimination in embedding space. Extensive experimental results show that the proposed method can not only achieve excellent robustness against adversarial attacks, but also derive a satisfactory  performance for out-of-distribution (OOD) detection. In particular, the proposed method achieves a classification accuracy of $32\%$ against PGD on CIFAR-10, and an F1 score of $0.937$ for OOD detection, where CIFAR-10 and LSUN are used as in-distribution and out-of-distribution data, respectively. In summary, the contribution of this study can be concluded as following:

\begin{itemize}
    \item A regularization, termed Dynamic Feature Aggregation, is proposed to compress the embedding space of the given model to improve its robustness against adversarial samples.

    \item The embedding codes with different classes are orthogonally processed for class-wise prediction, driving the model to learn distinguishable representations against trivial solution.

    \item The effectiveness of the proposed method against adversarial attacks is justified by theoretical proof. Compared to the zoo of Mixup, our method can more easily drive the model to meet the Lipschitz constraint.

    \item The proposed method, for the first time, constructs the connection between out-of-distribution detection and adversarial robustness. The compact embedding space leveraged by our method can not only defense the adversarial attacks, but also improve the performance of out-of-distribution detection.

\end{itemize}

\section{Related Work}
This study has connections to various areas, including adversarial attack and defense, zoo of Mixup and out-of-distribution detection. We thus briefly review the representative related works in this section.

\subsection{Adversarial Attack}
According to the accessibility of the model under attack, existing adversarial attack methods can be categorized into two protocols, \emph{i.e.,} white-box \cite{goodfellow2014explaining,athalye2018obfuscated,carlini2017towards,madry2017towards} and black-box \cite{ilyas2018prior,guo2019simple,brendel2017decision} settings. Since this work focuses on the defense against adversarial attacks, we mainly consider the worst case, \emph{i.e.,} white-box setting, where the attacker has the full knowledge of the model (\emph{e.g.,} parameters, gradient and architecture). In particular, three solid white-box attack methods, including FGSM, PGD and CW are employed in this paper. Fast gradient sign method (FGSM), proposed by Goodfellow \emph{et al.} \cite{goodfellow2014explaining}, generates adversarial examples with a single-step gradient update. Generally, FGSM is an attack with the low-level computation cost; however, its attack success rate may be unsatisfactory. 
In contrast, projected gradient descent (PGD) \cite{athalye2018obfuscated, madry2017towards} is a strong attack method, which can be regarded as an iterative variant of FGSM. The method starts from a random noise within the allowed norm ball, and then follows by multiple iterations of FGSM to generate adversarial perturbation. Apart from single/multi-step attacks, an optimization based method, called Carlini \& Wagner (CW) \cite{carlini2017towards} attack, is also considered in this paper. The CW method directly minimizes the distance between the benign and adversarial samples for 
attacks.

\subsection{Adversarial Attack Defense}
The adversarial attacks seriously threaten the application of CNN based models in security-critical scenarios. To address the problem, extensive studies have been proposed, which can be briefly categorized into training with or without the adversarial samples. In particular, Goodfellow \emph{et al.} \cite{goodfellow2014explaining} proposed the adversarial training, in which CNN based model is directly trained on adversarial samples. The adversarially trained model can withstand strong attacks, but leads to some new challenges, such as high computational cost \cite{shafahi2019adversarial} and over-fitting \cite{rice2020overfitting}. Although recent studies \cite{lee2020adversarial,shafahi2019adversarial,rice2020overfitting,bai2021recent} attempted to mitigate the issues, their performances are still unsatisfactory.
Therefore, defensing without using adversarial samples is gradually considered as a more promising research line. For example, the deep ensemble approach \cite{lakshminarayanan2016simple} creates an ensemble by training multiple models of the same architecture with different initializations. Since the adversarial attacks cannot simultaneously spoof multiple models effectively, the ensemble model yielded by deep ensembles naturally gains the excellent robustness against adversarial attacks. However, the method, which needs to train multiple models, causes an extremely high computational cost.
Another research line for attack defensing focuses on editing the input data to alleviate the influence caused by adversarial perturbations. These methods are generally based on randomized smoothing \cite{cohen2019certified}, random transformation \cite{liao2018defense} and image compression \cite{jia2019comdefend}. Under the white-box setting, these defense methods fail to achieve competitive performances, especially in the case that attack methods consider the combination of viewpoint shifts, noise, and other transformations \cite{athalye2018synthesizing}.

\subsection{Zoo of Mixup}
Since regularization based methods can simultaneously improve the generalization and robustness with neglectable extra computation costs, this field attracts an increasing attention from the community \cite{zhang2017mixup, guo2019mixup, yun2019cutmix}. To some extent, Mixup \cite{zhang2017mixup} is the first study that introduces sample interpolation strategy for the regularization of CNN based models. The virtual training sample, which is generated via the linear interpolation with pair-wise samples, smooths the network prediction. Following this direction, many variants were proposed by changing the form of interpolation. Manifold mixup \cite{verma2019manifold} generalized the Mixup to feature space. Guo \emph{et al.} \cite{guo2019mixup} proposed an adaptive Mixup by preventing the misleading generation caused by random mixing. Yun \emph{et al.} \cite{yun2019cutmix} proposed CutMix, which drew region-based interpolation between images rather than global mixing. In more recent studies, Puzzle Mix proposed by Kim \emph{et al.} \cite{kim2020puzzle} attempted to utilize the saliency information of each input for virtual sample generation.
Referring to the reported results of \cite{kim2020puzzle}, although many Mixup variants have been proposed, their improvements to model robustness against adversarial attacks are limited, compared to the vanilla Mixup.

In this paper, we revisit the interpolation-based strategy, and regard the mixed sample as a pivot for feature aggregation, which gives a new perspective for the application of mixing operation in the adversarial machine learning. For the first time, the model with interpolation based regularization can gain competitive robustness against strong attacks (\emph{e.g.,} PGD and CW) without using any adversarial samples.

\subsection{Out-of-Distribution Detection}
In addition to the adversarial attacks, the OOD sample is also a serious threat to the security of CNN based models. To deal with the problem, various OOD detection methods have been proposed \cite{hendrycks2016baseline, neal2018open, yoshihashi2019classification,liang2017enhancing,lee2018simple,liu2019large}. Liang \emph{et al.} \cite{liang2017enhancing} proposed ODIN, which adopted temperature scaling and pre-processing to enlarge the differences between in-distribution (ID) and OOD samples. Lee \emph{et al.} \cite{lee2018simple} employed Mahalanobis distance coupled with the pre-processing adopted in \cite{liang2017enhancing} for the identification of OOD samples. These two methods share a common drawback---their performances are sensitive to the hyper-parameters of pre-processing. Hence, they require a sophisticated fine-tuning using OOD data to achieve the satisfactory OOD detection performance.
However, in the real scenario, the OOD samples are difficult to acquire for training, which limits the potential application of these methods.

The methods, that relax the requirement for OOD samples are the more reliable choice for practical applications. Yoshihashi \emph{et al.} \cite{yoshihashi2019classification} thus proposed a OOD-sample-free pipeline to detect OOD samples by formulating an auxiliary reconstruction task. Neal \emph{et al.} \cite{neal2018open} introduced generative adversarial networks to OOD detection. However, for both reconstruction-based and generative OOD detection frameworks, their performances are sensitive to the quality of sample generation. Particularly, these methods cannot achieve the satisfactory OOD detection performance, dealing with the diverse data in limited scales.
More recently, Liu \emph{et al.} \cite{liu2019large} proposed an OLTR algorithm, which adopts memory features to limit the embedding space and enlarges the discrimination of OOD samples.  The capacity of the feature space is difficult to dynamically adapt to different situations and tasks. Such an issue makes the model performance sensitive to the hyper-parameter, \emph{i.e.,} memory size. In this paper, we propose a regularization, denoted as Dynamic Feature Aggregation, to mitigate the problem. Without constructing memory features, we draw a linear combination-based pivot as the center to aggregate features. By combining with a baseline OOD detection model \cite{zaeemzadeh2021out}, the proposed method significantly outperforms the memory feature bank based methods, and achieves the-state-of-the-art results on a publicly available benchmark \cite{lee2018simple,liang2017enhancing,liu2019large,zaeemzadeh2021out}.

\begin{figure*}[tbp]
    \centering
    \includegraphics[width=.88\textwidth]{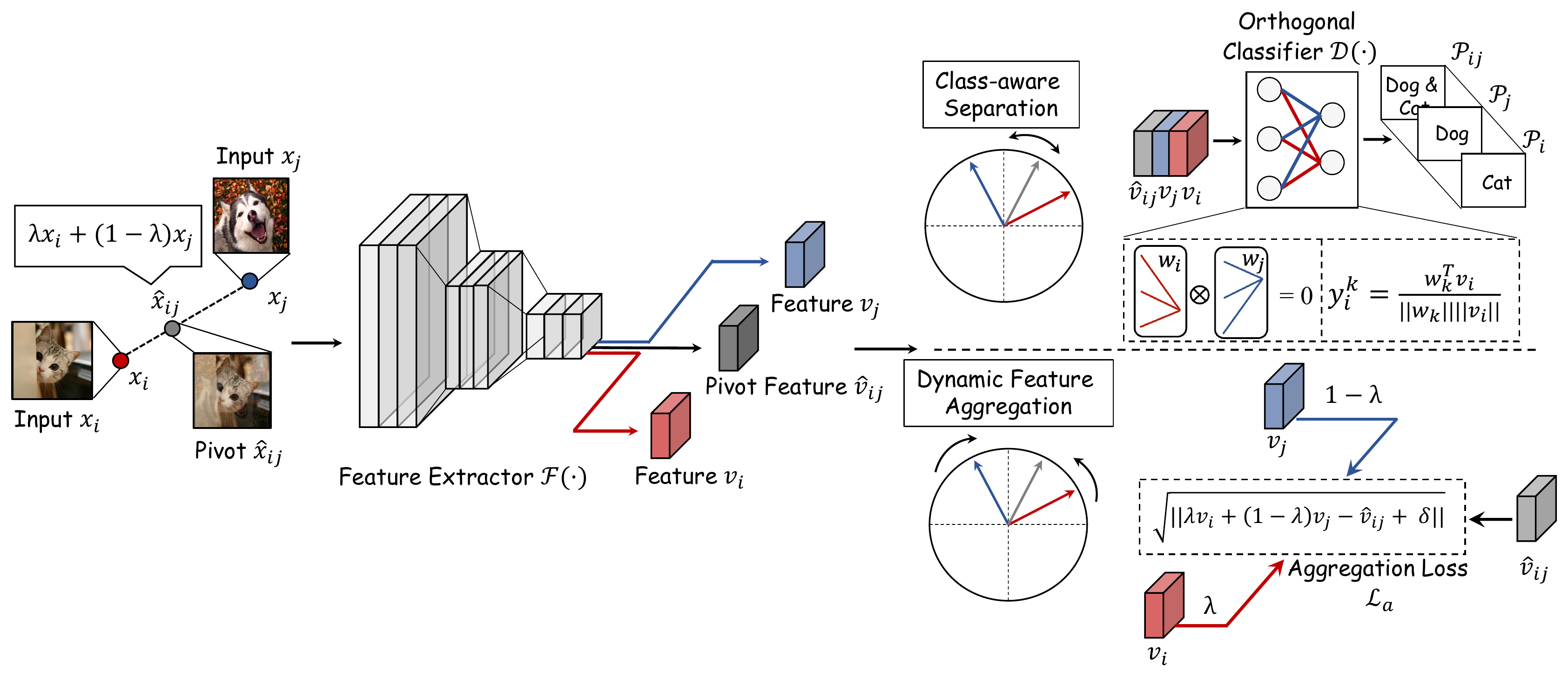}
    \caption{The pipeline of the proposed method. The CNN based model is trained through a dynamic manner: The proposed orthogonal classifier drives the model to learn distinguishable features to achieve class-aware separation. While, the proposed regularization, Dynamic Feature Aggregation, enforces the model to represent two given samples as similar as possible with their pivot, i,e. convex combination. The learning objective can be explained as compressing embedding space without sacrificing the performance of classification.}
    \label{fig:pipeline}
\end{figure*}

\section{Proposed Method}
\label{sec:method}
As the overview shown in Fig. \ref{fig:pipeline}, the proposed method drives the CNN based model to play a dynamic game. In particular, the proposed regularization, termed Dynamic Feature Aggregation, compresses the embedding space by minimizing the distance between two given samples and their convex combinations. While, an orthogonal classifier is employed to guide the model to represent images within class-aware separation. By taking such a dynamic game as the learning objective, the model can achieve a compact-but-distinguishable embedding space, which benefits its robustness against adversarial attacks and OOD samples.

\subsection{Dynamic Feature Aggregation}
\label{sec:DFA}
According to the pipeline shown in Fig. \ref{fig:pipeline}, a feature extractor $\mathcal{F}(\cdot)$ with multiple CNN layers is adopted to encode the input image $x \in \mathcal{X}$ into an embedding code $v \in \mathcal{V}$. Since the proposed method aims to improve model robustness by aggregating features within the embedding space $\mathcal{V}$, we first localize the center for aggregation. Given two different samples $x_i$ and $x_j$, their convex combination can be defined as:
\begin{align}
    \hat{x}_{ij} = \Lambda_{\lambda}(x_i,x_j) = \lambda x_i + (1 - \lambda) x_j
\end{align}
where $\Lambda_{\lambda}(\cdot,\cdot)$ is the function of linear combination with a hyper-parameter sampled from Beta distribution, \emph{i.e.,} $\lambda \in \mathbb{B}(\alpha,\alpha)$. By drawing $\{ x_i, x_j, \hat{x}_{ij} \}$ as inputs, the features $\{v_i, v_j, \hat{v}_{ij}\}$ are respectively embedded via $\mathcal{F(\cdot)}$. Since $\hat{v}_{ij}$ is calculated through the combination of $x_i$ and $x_j$,
it contains the common patterns of $x_i$ and $x_j$, which leads to an ideal center for the aggregation of $v_i$ and $v_j$. In this paper, mean squared error (MSE) is adopted as the distance evaluation metric on $\mathcal{V}$; hence, the feature aggregation is achieved by minimizing the aggregation loss $\mathcal{L}_a$, defined as:
\begin{align}
\label{eq:la}
    \mathcal{L}_a \!\!=\!\!\!\!\!\! \iint\limits_{x_i,x_j \in \mathcal{X}}\!\!\!\!\! \sqrt{ || \lambda \mathcal{F}(x_i) \!+\! (1-\lambda) \mathcal{F}(x_j)\! -  \!\hat{v}_{ij} \!+\! \sigma|| } \; dx
\end{align}
where $\sigma$ is a noise term sampled from Gaussian noise to prevent over-fitting. By pulling the distance between $\hat{v}_{ij}$ and $\{v_i,v_j\}$ in such a re-weighting strategy, a more compact embedding space can be obtained. Therefore, there may be limited embedding spaces left for the generation of adversarial attacks.

\subsection{Class-aware Separation}
As a strong constraint for the classification task, Dynamic Feature Aggregation brings a trivial solution to $\mathcal{F}(\cdot)$, \emph{i.e.,} each $x \in \mathcal{X}$ is encoded into a fixed $v$. Inspired by \cite{zaeemzadeh2021out}, we design an orthogonal classifier, $\mathcal{D}(\cdot)$, to alleviate the problem by ensuring the class-aware separation on $\mathcal{V}$. Concretely, $\mathcal{D}(\cdot)$ consists of several weight vectors $w_k \in \mathcal{W}$, where $w_k$ refers to the $k$-th category. Unlike vanilla CNN based model, we replace the last fully-connected layer by $\mathcal{D}(\cdot)$. Therefore, $k$-class prediction score of $x_i$ (\emph{i.e.,} $y^k$) can be given as:
\begin{align}
    y^k_i = \frac{w_k^T v_i}{||w_k||\; ||v_i||}.
\end{align}
The $k$-th class probability $p^k_i$ of $x_i$ can be accordingly calculated via a softmax layer:
\begin{align}
    p^k_i = \frac{exp{(y^k_i})}{\sum\limits_{1 \leq l \leq n} exp({y^l_i})}
\end{align}
where the set $\mathcal{P}_i = \mathcal{D}(v_i) = \{ p^k_i| 1 \leq k \leq n \}$ forms the final output of the CNN based model for an $n$-class recognition task.
By removing the bias and activation function in the last layer, CNN based model maps $x$ into the allowed norm ball space, which can effectively compress the embedding space and alleviate the problem of trivial solution. To further strengthen the class-aware separation in the embedding space, we then introduce the orthogonal constraint to initialize $\mathcal{W}$, which can be written as:
\begin{align}
    \label{eq:or}
    \prod_{w_k,w_l \in \mathcal{W},k\neq l} w_k^T w_l = 0.
\end{align}
Since the orthogonal constraint and Dynamic Feature Aggregation are both strong constraints for a CNN based model, training with both terms might cause sharp gradient updates, which may degrade the classification performance. We thus propose to freeze $\mathcal{W}$ when training $\mathcal{F}(\cdot)$ to stabilize the representation in embedding space.  Note that the training process is not iterative. $\mathcal{W}$ is frozen after initialization.

To sum up, the learning objective of the proposed method can be finally concluded as:
\begin{equation}
\label{eq:lc}
    \begin{aligned}
        \mathcal{L}_c =  \iint\limits_{x_i,x_j \in \mathcal{X}}  \Lambda_{\lambda}(\widetilde{y}_i,\widetilde{y}_j) & log(\mathcal{D}(\mathcal{F}(\Lambda_{\lambda}(x_i,x_j)))) \\
        + \Lambda_{\lambda}(\widetilde{y}_i,\widetilde{y}_j) & log(\mathcal{D}(\Lambda_{\lambda}(v_i,v_j))\; dx
    \end{aligned}
\end{equation}
\begin{align}
    \mathcal{L}_{t} =   \mathcal{L}_c + \mathcal{L}_a
\end{align}
where $\widetilde{y}_i$ refers to the category-wise label of $x_i$; $\mathcal{L}_c$ is the mixing-based cross-entropy loss the same as \cite{zhang2017mixup}; and $\mathcal{L}_{t}$ represents the final loss function of the proposed method. For clarity, the proposed method is summarized in Alg.~\ref{algo:2}.
\begin{algorithm}[!tb]
	\caption{Robust Representation via Dynamical Feature Aggregation}\label{algo:2}
	\begin{algorithmic}
    \Require\\
     Feature Extractor  $\mathcal{F}(\cdot)$; Orthogonal Classifier $\mathcal{D}(\cdot)$;
    \Ensure \State Trained  $\mathcal{F}(\cdot)$;
  \end{algorithmic}
  \begin{algorithmic}[1]
    \State Initialize $\mathcal{D}(\cdot)$ through Eq. (\ref{eq:or});
    \While{ $\mathcal{F}(\cdot)$ has not converged}
        \State Sample two training images $x_i,x_j \sim \mathcal{X}$;
            \State Sample $\lambda_0$ from $\mathbb{B}(\alpha,\alpha)$;
            \State $\hat{x}_{ij} \leftarrow \Lambda_{\lambda_0}(x_i,x_j)$;
            \State $ v_i,v_j,\hat{v}_{ij} \leftarrow \mathcal{F}(x_i), \mathcal{F}(x_j), \mathcal{F}(\hat{x}_{ij})$;
            \State Calculate $\mathcal{L}_a$ and $\mathcal{L}_c$ by Eq. (\ref{eq:la}) and Eq. (\ref{eq:lc}), respectively;
            \State $\mathcal{L}_{t} \leftarrow \mathcal{L}_c + \mathcal{L}_a$;
            \State Update $\mathcal{F}(\cdot)$ by minimizing $\mathcal{L}_{t}$;
 \EndWhile
\State Return $\mathcal{F}(\cdot)$;
	\end{algorithmic}
\end{algorithm}

\section{Theoretical Analysis for the Boosted Robustness over the Zoo of Mixup}
To further investigate the superiority of our method over the other mixing-based approaches, we theoretically analyze the proposed Dynamic Feature Aggregation and show it is a precise and universal solution for Lipschitz continuity, while other mixing-based methods are not.

\vspace{3mm}
\noindent \textbf{Preliminary.} In the proposed method, the feature extractor $\mathcal{F}(\cdot)$ connects the input space $\mathcal{X}$ and the embedding space $\mathcal{V}$. Given two evaluation metrics $\mathbb{D}_x(\cdot,\cdot)$ and $\mathbb{D}_v(\cdot,\cdot)$ defined on $\mathcal{X}$ and $\mathcal{V}$, respectively, $\mathcal{F}(\cdot)$ fulfills Lipschitz continuity, if a real constant $K$ is existed to ensure all $x_i,x_j \in \mathcal{X}$ meet the following condition:
\begin{align}
\label{eq:lip}
    K \mathbb{D}_x(x_i,x_j) \geq  \mathbb{D}_v(\mathcal{F}(x_i),\mathcal{F}(x_j)).
\end{align}

\vspace{3mm}
\noindent \textbf{Proposition.} Based on the analysis in \cite{verma2019manifold,cisse2017parseval}, a flat embedding space, especially with Lipschitz continuity, is an ideal solution against adversarial attack. Hence, the effectiveness of the proposed method can be justified by proving the equivalence between Dynamic Feature Aggregation and $K$-Lipschitz continuity.

\vspace{3mm}
\noindent \textbf{Theorem.} Towards any $K$ of Lipschitz continuity, Dynamic Feature Aggregation is a precise and universal solution.

\vspace{3mm}
\noindent \textbf{Hypothesis.} $\mathbb{D}_x(\cdot,\cdot)$ is a linear metric and $\mathbb{D}_v(\cdot,\cdot)$ performs as 2-norm distance:
\begin{align}
    \mathbb{D}_x(x_1,x_2) \!+\! \mathbb{D}_x(x_3,x_4) &\!=\! \mathbb{D}_x(x_1,x_4) + \mathbb{D}_x(x_2,x_3), \\
    \mathbb{D}_v(v_1,v_2) &\!=\! ||v_1^T v_1 \!-\! 2v_1^T v_2 \!+\! v_2^T v_2||.
\end{align}

\vspace{3mm}
\noindent \textbf{Proof.} Given $x_i$ and $x_j$ sampled from $\mathcal{X}$, their convex combination based pivot $\hat{x}_{ij}$ can be obtained via $\Lambda_{\lambda}(x_i,x_j)$. Since $\hat{x}_{ij}$ can be regarded as a sample in $\mathcal{X}$, we can transform Lipschitz continuity (Eq. (\ref{eq:lip})) to
\begin{equation}
\begin{aligned}
        &K\mathbb{D}_x(\hat{x}_{ij},\hat{x}_{ij}) =K  \mathbb{D}_x(\hat{x}_{ij}, \Lambda_{\lambda}(x_i,x_j)) \\
         &                                       =K( \Lambda_{\lambda}(\mathbb{D}_x(\hat{x}_{ij},x_i), \mathbb{D}_x(\hat{x}_{ij},x_j))-  \mathbb{D}_x(\hat{x}_{ij}, \mathcal{O}) \\
          &                                       -\! \mathbb{D}_x(\lambda x_i,(1-\lambda)x_j)
          \!+\! \mathbb{D}_x(\lambda x_i,\mathcal{O}) \!+\! \mathbb{D}_x((1-\lambda)x_j,\mathcal{O}) )\\
          & \geq  \Lambda_{\lambda}(\mathbb{D}_v(\hat{v}_{ij},v_i), \mathbb{D}_v(\hat{v}_{ij},v_j))-  \mathbb{D}_v(\hat{v}_{ij}, \mathcal{O}) \\
          &                                       - \mathbb{D}_v(\lambda v_i,(1-\lambda)v_j) + \mathbb{D}_v(\lambda v_i,\mathcal{O}) + \mathbb{D}_v((1-\lambda)v_j,\mathcal{O}) \\
          &= \mathbb{D}_v(\hat{v}_{ij}, \Lambda_{\lambda}(v_i,v_j))
\end{aligned}
\end{equation}
where $\mathcal{O}$ refers to the matrix zero. Based on the hypothesis of $\mathbb{D}_v(\cdot,\cdot)$ as a 2-norm distance, $\mathbb{D}_v(\cdot,\cdot)$ should be no less than 0. Hence, we can get the lower and upper bound of $\mathbb{D}_v(\hat{v}_{ij}, \Lambda_{\lambda}(v_i,v_j))$ within Lipschitz continuity:
\begin{align}
    0 \leq \mathbb{D}_v(\hat{v}_{ij}, \Lambda_{\lambda}(v_i,v_j)) \leq K\mathbb{D}_x(\hat{x}_{ij},\hat{x}_{ij}) = 0.
\end{align}
Therefore, $\mathbb{D}_v(\hat{v}_{ij}, \Lambda_{\lambda}(v_i,v_j))$ should be zero. Based on Eq. (\ref{eq:la}), the optimal result of Dynamic Feature Aggregation is identical to $\mathbb{D}_v(\hat{v}_{ij}, \Lambda_{\lambda}(v_i,v_j))$. Therefore, $K$-Lipschitz continuity can be ensured by minimizing Dynamic Feature Aggregation.

\vspace{3mm}
\noindent \textbf{Superiority to the Zoo of Mixup.}
Existing Mixup based methods interpolated pair-wise inputs and their labels. Hence, the formal learning objective of such methods can be concluded as:
\begin{align}
     \iint\limits_{x_i,x_j \in \mathcal{X}} \Lambda_{\lambda}(\widetilde{y}_i,\widetilde{y}_j) & log(\mathcal{D}(\hat{v}_{ij})\; dx
\end{align}
which drives $\mathcal{F}(\cdot)$ and $\mathcal{D}(\cdot)$ to minimize the distance between $\mathcal{D}(\hat{v}_{ij})$ and $\Lambda_{\lambda}(\widetilde{y_i},\widetilde{y_j})$. Based on the \textbf{Proof}, the Lipschitz continuity of $\mathcal{F}(\cdot)$ and $\mathcal{D}(\cdot)$ can be achieved when
\begin{align}
\mathcal{D}(\hat{v}_{ij}) = \Lambda_{\lambda}({\mathcal{P}_i},\mathcal{P}_j),
\end{align}
where $P_i = \mathcal{D}(v_i)$. This indicates, if and only if $\Lambda_{\lambda}({\mathcal{P}_i},\mathcal{P}_j)$ equals to $\Lambda_{\lambda}(\widetilde{y_i},\widetilde{y_j})$, the zoo of Mixup can achieve the optimal result for Lipschitz continuity. However, in the real scenario, there is a non-negligible gap between the two terms. For example, model trained on CIFAR-100 can only achieve $\sim80\%$ accuracy. Hence, it accordingly verifies
the superiority of the proposed method, which can directly fulfill Lipschitz constraint, to the zoo of Mixup.

\section{Class-aware OOD Detection using Dynamic Feature Aggregation}
As a regularization, Dynamic Feature Aggregation can be easily integrated to the other proposed methods. To clarify the contribution of Dynamic Feature Aggregation for OOD detection, we combine the proposed method with the existing OOD detection model of \cite{zaeemzadeh2021out}. Given all training samples $x \in \mathcal{X}$ as inputs, we can obtain the corresponding representation $v \in \mathcal{V}$ via the trained $\mathcal{F}(\cdot)$. Then, the representative representation $v^*_k$ of the $k$-th class can be obtained by computing the first singular vectors of $\{x_i \in \mathcal{X}\;|\; \mathop{\arg\max}\widetilde{y_i} = k \}$. Taking a test sample $x_t$ as an example, the OOD probability $\phi_t$ can be accordingly calculated.
\begin{align}
    \phi_t = \min_k \text{arccos}(\frac{|\mathcal{F}^T(x_t) v_k^*|}{||\mathcal{F}(x_t)||})
\end{align}
where $x_t$ is categorized as an OOD sample if $\phi_t$ is larger than a predefined threshold $\phi^*$.

\section{Experimental Results and Analysis}
To evaluate the performance of the proposed method, extensive experiments are conducted on different tasks, including adversarial attack defensing and OOD detection. In this section, we first present the information of datasets and the implementation details. Then, the robustness of the proposed method against adversarial attack is empirically validated by comparing with the zoo of Mixup. Finally, the effectiveness of the proposed method for OOD detection is evaluated on a publicly available benchmarking dataset.

\subsection{Experimental Settings}\label{adrobust}
In this section, we introduce the datasets and implementation details for adversarial attack defensing and OOD detection tasks, respectively.

\begin{table*}[!htbp]
\centering
\caption{Accuracy (\%) on CIFAR-10/100 based on WRN-28-10 trained with various methods. M.-Mixup. stands for Manifold Mixup. $\dagger$ refers to the reproduced results by using the publicly-available code. \textbf{Bold} denotes the best result.}
\label{tab:cifar10}
\setlength\tabcolsep{6pt}
\resizebox{.98\textwidth}{!}{
\begin{tabular}{c|c|ccccc|c}
\hline
CIFAR-10/100                      & Clean          & \begin{tabular}[c]{@{}c@{}}FGSM \\ (8/255)\end{tabular} & \begin{tabular}[c]{@{}c@{}}PGD-8\\ (4/255)\end{tabular} & \begin{tabular}[c]{@{}c@{}}PGD-16\\ (4/255)\end{tabular} & \begin{tabular}[c]{@{}c@{}}CW-100\\ (c=0.01)\end{tabular} & \begin{tabular}[c]{@{}c@{}}CW-100\\ (c=0.05)\end{tabular} & \textbf{Mean $\pm$ S.d.}            \\ \hline
Baseline reported in \cite{zagoruyko2016wide}      & 96.11/81.15          & -                                                       & -                                                       & -                                                        & -                                                         & -                                                         & -                          \\
Baseline$\dagger$                     & 96.28/80.82          & 38.03/11.71                                                   & 0.92/0.79                                                    & 0.28/0.42                                                     & 11.1/4.42                                                      & 0.39/0.23                                                      & 10.14 $\pm$ 14.53/3.51 $\pm$4.38          \\\hline
Mixup reported in \cite{zhang2017mixup}           & 97.08/81.11          & -                                                       & -                                                       & -                                                        & -                                                         & -                                                         & -                          \\
Mixup$\dagger$                        & 97.01/82.75          & 60.17/27.34                                                   & 3.97/0.28                                                    & 1.16/0.11                                                     & 30.32/4.83                                                     & 2.36/0.28                                                      & 19.60 $\pm$ 22.99/6.57 $\pm$ 10.54          \\\hline
M.-Mixup reported in \cite{verma2019manifold}  & \textbf{97.45}/81.96          & -                                                       & -                                                       & -                                                        & -                                                         & -                                                         & -                          \\
M.-Mixup$\dagger$               & 97.12/\textbf{83.94} & 59.32/\textbf{29.73}                                                   & 7.97/1.19                                                    & 2.97/0.49                                                     & 51.47/10.75                                                     & 11.12/0.77                                                     & 26.57 $\pm$ 23.80/8.59 $\pm$11.25          \\\hline
\rowcolor[HTML]{EFEFEF}
Ours                         & 96.80/82.02          & \textbf{74.18}/24.28                                          & \textbf{32.12}/\textbf{8.22}                                          & \textbf{22.12}/\textbf{7.40}                                           & \textbf{81.39}/\textbf{42.02}                                            & \textbf{74.72}/\textbf{26.18}                                            & \textbf{56.91 $\pm$ 24.66/21.62 $\pm$ 12.85} \\ \hline\hline
Adversarial Training \cite{shafahi2019adversarial} $\dagger$        & 92.18/71.00          & 48.99/23.12                                                   & 66.77/36.22                                                   & 66.42/35.78                                                    & 87.14/61.26                                                     & 67.18/50.79                                                     & 67.30 $\pm$ 12.08/41.43 $\pm$ 13.23          \\ \hline
\end{tabular}
}
\vspace{-3mm}
\end{table*}
\begin{table*}[]
\centering
\caption{Accuracy (\%) on Tiny Imagenet based on PreActResNet18 trained with various methods. M.-Mixup. stands for Manifold Mixup. $\dagger$ refers to the reproduced results by using the publicly-available code. \textbf{Bold} denotes the best result.}
\label{tab:imagenet}
\setlength\tabcolsep{20pt}
\Large
\resizebox{.98\textwidth}{!}{
\begin{tabular}{c|c|ccccc|c}
\hline
Tiny-ImageNet        & Clean          & \multicolumn{1}{c}{\begin{tabular}[c]{@{}c@{}}FGSM\\ (8/255)\end{tabular}} & \multicolumn{1}{c}{\begin{tabular}[c]{@{}c@{}}PGD-8\\ (4/255)\end{tabular}} & \multicolumn{1}{c}{\begin{tabular}[c]{@{}c@{}}PGD-16\\ (4/255)\end{tabular}} & \multicolumn{1}{c}{\begin{tabular}[c]{@{}c@{}}CW-100\\ (c=0.01)\end{tabular}} & \begin{tabular}[c]{@{}c@{}}CW-100\\ (c=0.05)\end{tabular} & \multicolumn{1}{c}{\textbf{Mean $\pm$ S.d.}} \\ \hline
Baseline \cite{he2016identity} & 55.52          & -                                                                           & -                                                                            & -                                                                             & -                                                                              & -                                                         & -                                             \\
Baseline$\dagger$              & 58.62          & 4.26                                                                        & 0.81                                                                         & 0.60                                                                          & 27.92                                                                          & 7.52                                                      & 8.22 $\pm$ 10.17                              \\ \hline
Mixup \cite{zhang2017mixup} & 56.47          & -                                                                           & -                                                                            & -                                                                             & -                                                                              & -                                                         & -                                             \\
Mixup$\dagger$                 & 61.03          & 4.23                                                                        & 0.98                                                                         & 0.77                                                                          & 29.13                                                                          & 15.41                                                     & 10.10 $\pm$ 10.91                             \\ \hline
M.-Mixup \cite{verma2019manifold}& 58.70          & -                                                                           & -                                                                            & -                                                                             & -                                                                              & -                                                         & -                                             \\
M.-Mixup $\dagger$             & \textbf{61.97} & 3.04                                                                        & 0.82                                                                         & 0.59                                                                          & 29.69                                                                          & 16.86                                                     & 10.20 $\pm$ 11.45                             \\ \hline
\rowcolor[HTML]{EFEFEF}
Ours                 & 60.59          & \textbf{7.10}                                                               & \textbf{4.66}                                                                & \textbf{4.98}                                                                 & \textbf{35.93}                                                                 & \textbf{34.22}                                            & \textbf{17.38 $\pm$ 14.48}                    \\ \hline
Adversarial Training \cite{shafahi2019adversarial} $\dagger$  & 55.67          & 10.99                                                                       & 20.76                                                                        & 20.38                                                                         & 45.77                                                                          & 45.65                                                     & 28.71 $\pm$ 14.32                             \\ \hline
\end{tabular}
}
\end{table*}

\vspace{3mm}
\noindent \textbf{a) Adversarial Robustness.} The robustness of the proposed method against adversarial attacks is evaluated on CIFAR-10, CIFAR100 \cite{krizhevsky2009learning} and Tiny ImageNet. The WideResNet \cite{zagoruyko2016wide} with depth of 28 and width of 10 (WRN-28-10) is adopted as the backbone for CIFAR-10/100. While, for the Tiny-ImageNet \cite{deng2009imagenet}, the backbone is set as PreActResNet18 \cite{he2016identity}. For data augmentation, we employ horizontal flipping and cropping from the image padded by four pixel on each side in this experiment. To guarantee the fairness of performance comparison, all the experiments are conducted under the same training protocol. In particular, the models are trained using SGD with a weight decay of 0.0005 and a momentum of 0.9. Models are observed to converge after 200 epochs of training. The list of learning rate is set to [0.1, 0.02, 0.004, 0.0008], in which the learning rate decreases to the next after every 60 training epochs. The noise term $\sigma$ is set to 0.05 for CIFAR-10 and 0.005 for CIFAR-100, respectively.

\vspace{1mm}
{\itshape Attack Methods.} To evaluate the robustness against adversarial attacks, three popular adversarial attack methods, including FGSM \cite{goodfellow2014explaining}, PGD \cite{athalye2018obfuscated,madry2017towards} and CW \cite{carlini2017towards}, are involved in this study. The perturbation budget is set to 8/255 and 4/255 under $l_\infty$ norm distance for single- and multi-step attacks. PGD-$K$ denotes a $K$-step attack with a step size of 2/255. For CW, two cases are taken into account, in which the steps are both set to 100 steps and $c$ is set to 0.01 and 0.05, respectively.

\vspace{1mm}
{\itshape Benchmarking Methods and Evaluation Criterion.} Two interpolation-based methods, including Mixup \cite{zhang2017mixup} and Manifold-Mixup \cite{verma2019manifold}, are involved for comparison in this study. Although there are recent papers proposing new ways to mix samples in the input space \cite{kim2020puzzle,guo2019mixup,yun2019cutmix}, they do not achieve significant improvements over Mixup or Manifold-Mixup, especially against adversarial attacks \cite{kim2020puzzle}. Therefore, Mixup and Manifold-Mixup remain the most relevant competing methods among the zoo of Mixup. Note that our mixing strategy is based on Manifold-Mixup, which performs as a solid baseline to validate the effectiveness of Dynamic Feature Aggregation. For a more comprehensive analysis of the proposed method, an effective adversarial training, free-AT \cite{shafahi2019adversarial}, is also included for reference as the upper bound. The evaluation metric is the classification accuracy on the whole test set.

\vspace{3mm}
\noindent \textbf{b) Out-of-Distribution Detection.} In the OOD detection scenario, the training set of CIFAR-10 \cite{krizhevsky2009learning} is adopted as the in-distribution data, and the test set of CIFAR-10 refers to the positive samples for OOD detection. Similar to the prior works \cite{zaeemzadeh2021out,lee2018simple,liang2017enhancing,liu2019large}, the OOD datasets include Tiny-ImageNet \cite{deng2009imagenet} and LSUN \cite{yu2015lsun}. Tiny-ImageNet (a subset of ImageNet \cite{deng2009imagenet}) consists of 10,000 test images with a size of 36 $\times$ 36 pixels, which can be categorized to 200 classes. LSUN \cite{yu2015lsun} consists of 10,000 test samples from 10 different scene groups. Since the image size of Tiny-ImageNet and LSUN are not identical with that of CIFAR-10, two downsampling strategy (crop (C) and resize (R)) are adopted for image size unification, 
following the protocol of \cite{liang2017enhancing,vyas2018out,zaeemzadeh2021out}.
Therefore, we have four OOD test datasets, \emph{i.e.,} TIN-C, TIN-R, LSUN-C and LSUN-R. The training protocol and backbone for OOD detection is identical to Sec.~\ref{adrobust} a.

\vspace{1mm}
{\itshape Benchmarking Methods and Evaluation Criterion.} For the competing methods, Softmax Pred. \cite{hendrycks2016baseline}, Counterfactual \cite{neal2018open}, CROSR \cite{yoshihashi2019classification}, OLTR \cite{liu2019large} and Union of 1D Subspaces \cite{zaeemzadeh2021out} are included. We also exploit the solutions using Monte Carlo sampling or OOD samples \cite{liang2017enhancing,lee2018simple} as the references for competing methods. To some extent, these methods can be seen as the upper bound for OOD detection regardless of time-consumption and over-fitting.
For example, Monte Carlo sampling \cite{maddox2019simple,gal2016dropout} could generally yield improvements to most of current OOD methods with huge extra
computational costs. The evaluation metric for OOD detection is the F1 score---the maximum score over all possible threshold $\phi^*$.

\subsection{Performance against Adversarial Attacks}
\label{sec:adversarial}
To evaluate the adversarial robustness of the proposed method, we compare it with Mixup\cite{zhang2017mixup} and Manifold Mixup \cite{verma2019manifold} (denoted as M.-Mixup) and present the results in Table \ref{tab:cifar10} and Table \ref{tab:imagenet}. The model trained on clean data without using any interpolation-based augmentation methods is adopted as the baseline. As existing benchmarks only adopt FGSM to evaluate the robustness, we re-implement Mixup and Manifold-Mixup into a more comprehensive benchmark with various adversarial attacks. As listed in Table \ref{tab:cifar10}, the proposed method significantly outperforms the competing methods. The average classification accuracy of the proposed method can reach to $56.91\%$ on CIFAR-10, which surpasses Mixup and Manifold Mixup by margins of $37.31\%$ and $30.34\%$, respectively. Furthermore, the existing mixing based methods are observed to be vulnerable to the strong attacks, \emph{e.g.,} PGD.
In contrast, the proposed method yields a surprising improvement to the robustness against PGD.
Particularly, the proposed method achieves an accuracy of $32.12\%$ against PGD-8, while the accuracy of Mixup and Manifold Mixup is only $3.97\%$ and $7.97\%$, respectively. In terms of model generalization, an improvement of $+0.69\%$ on clean data (\emph{i.e.,} without adversarial attacks) is achieved by the proposed method upon the baseline. In other words, our Dynamic Feature Aggregation can improve the robustness of the model without sacrificing the generalization.

Furthermore, the proposed model is also tested on a larger-scale dataset, \emph{i.e.,} Tiny-ImageNet. As listed in Table \ref{tab:imagenet}, the proposed approach can achieve the superior performance against different kinds of adversarial attacks, compared to the other interpolation based methods. Concretely, by using our dynamic feature aggregation, the model can achieve an average accuracy of $17.38\%$, surpassing Manifold Mixup by a margin of $\sim$7\%.

\begin{table}[!htbp]
\centering
\caption{Accuracy (\%) on CIFAR-10 based on WRN-28-10 trained with the various methods with orthogonal classifier (Orth.).}
\label{tab:ablation study}
\Huge
\resizebox{.48\textwidth}{!}{
\begin{tabular}{c|ccccc|c}
\hline
CIFAR10                               & \begin{tabular}[c]{@{}c@{}}FGSM \\ (8/255)\end{tabular} & \begin{tabular}[c]{@{}c@{}}PGD-8\\ (4/255)\end{tabular} & \begin{tabular}[c]{@{}c@{}}PGD-16\\ (4/255)\end{tabular} & \begin{tabular}[c]{@{}c@{}}CW-100\\ (c=0.01)\end{tabular} & \begin{tabular}[c]{@{}c@{}}CW-100\\ (c=0.05)\end{tabular} & Mean $\pm$ S.d.            \\ \hline
Baseline                              & 38.03                                                   & 0.92                                                    & 0.28                                                     & 11.1                                                      & 0.39                                                      & 10.14 $\pm$ 14.53          \\\hline
Mixup  & 60.17 & 3.97 & 1.16 & 30.32 & 2.36 & 19.60 $\pm$ 22.99\\
Orth. + Mixup                    & 44.80                                                   & 3.99                                                    & 2.66                                                     & 71.12                                                     & 49.47                                                     & 34.41 $\pm$ 26.89          \\ \hline
M.-Mixup & 59.32 & 7.97 & 2.97 & 51.47 & 11.12 & 26.57 $\pm$ 23.80\\
Orth. + M.-Mixup           & 38.76                                                   & 5.77                                                    & 4.38                                                     & 69.08                                                     & 53.98                                                     & 34.39 $\pm$ 25.79          \\ \hline
\rowcolor[HTML]{EFEFEF}
Ours                                  & \textbf{74.18}                                          & \textbf{32.12}                                          & \textbf{22.12}                                           & \textbf{81.39}                                            & \textbf{74.72}                                            & \textbf{56.91 $\pm$ 24.66} \\ \hline
\end{tabular}
}
\end{table}
\begin{figure}[!htb]
    \centering
    \includegraphics[width=.48\textwidth]{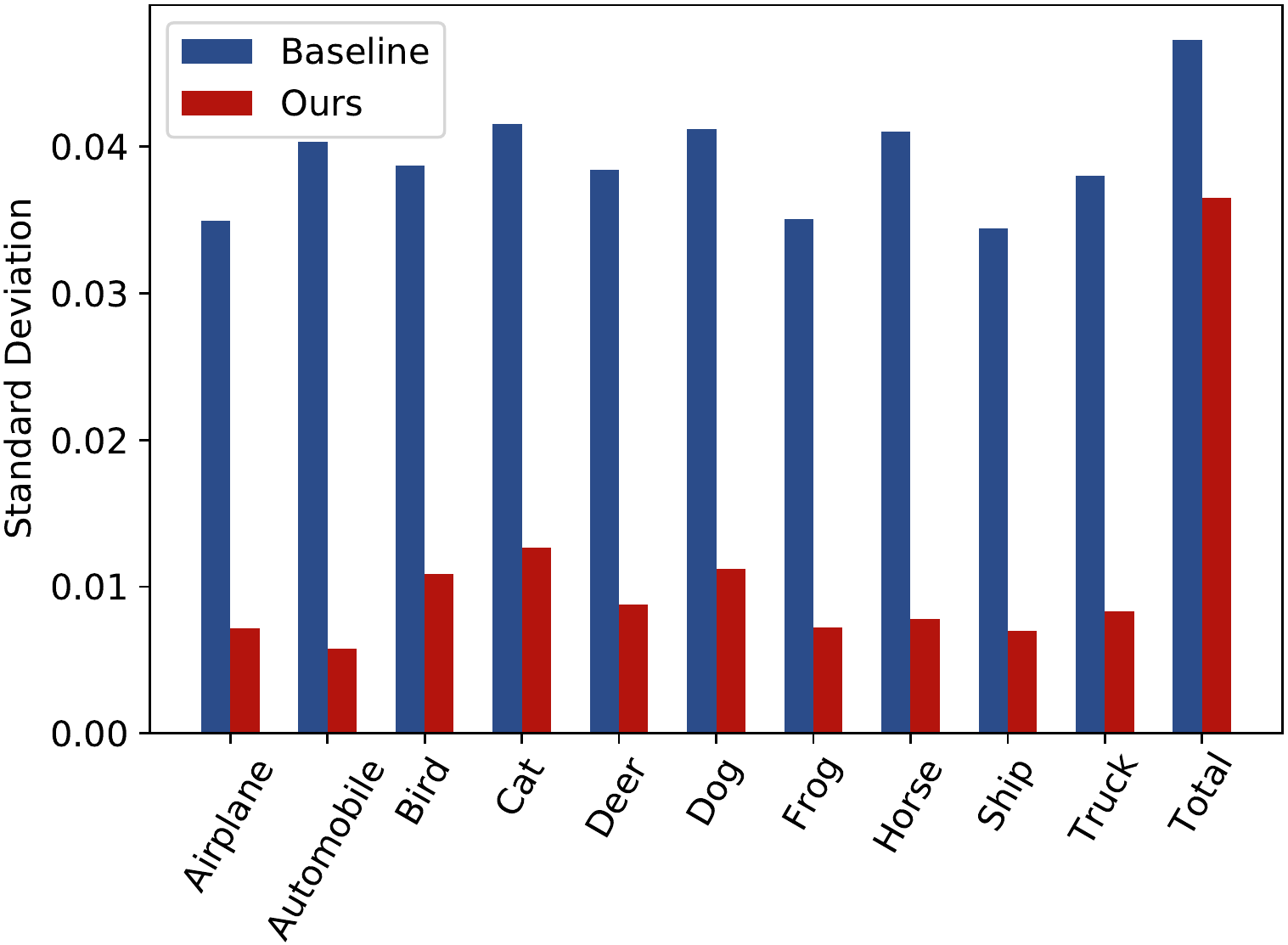}
    \caption{Compactness (\emph{i.e.,} standard deviation) of the embedding clusters on CIFAR-10. The `Total' is the compactness of the whole embedding space.}
    \label{fig:results}
    \vspace{-3mm}
\end{figure}

\vspace{3mm}
\noindent {\bf Ablation Study.} To quantify the contribution of Dynamic Feature Aggregation and orthogonal classifier,
we combine the orthogonal classifier with Mixup and Manifold Mixup for comparison. The evaluation results are listed in Table \ref{tab:ablation study}. Although orthogonal classifier improves the robustness of the corresponding benchmarking methods, there is no a significant improvement on the defense of PGD-8/16. Only with our Dynamic Feature Aggregation, the CNN can achieve a higher accuracy of $32.12\%$ under the PGD-8 attack, which validates the importance of Dynamic Feature Aggregation for improving model robustness against strong attacks. The excellent robustness of our Dynamic Feature Aggregation is due to its compact embedding space. To validate this claim, we also analyze the cluster compactness (\emph{i.e.,} standard deviation of the cluster) of each class in the feature space on CIFAR-10, which is presented in Fig. \ref{fig:results}. It can be observed that the class-wise standard deviation of our Dynamic Feature Aggregation is much lower than that of the baseline. The entry of `Total' measures the compactness of the whole embedding space. Our method is observed to compact the whole embedding space to a lower extent, compared to the class-wise clusters, to maintain the model generalization on clean data.


\vspace{3mm}
\noindent {\bf Evaluation on Hyper-parameters.} Here, we analyze the influence caused by different values of hyper-parameters, including $\sigma$ for the noise term and $\alpha$ for Beta Distribution. As listed in Table \ref{tab:noise}, noises on multiple levels $\sigma$=[0.1, 0.05, 0.01, 0.005, 0.001] are considered for the grid search. The best performance is achieved as $\sigma$=0.05.
In terms of $\alpha$, we conduct two settings for comparison, including $\alpha$=1.0 and 2.0. The model trained with $\alpha$=1.0 is observed to outperform the one with $\alpha$=2.0 (\emph{i.e.,} $56.91\%$ \emph{vs.} $55.04\%$).

\begin{table}[]
\centering
\caption{Accuracy (\%) on CIFAR-10 based on WRN-28-10 trained with the proposed method under various noise terms $\sigma$.}
\label{tab:noise}
\Huge
\resizebox{.48\textwidth}{!}{
\begin{tabular}{c|ccccc|c}
\hline
Noise  & \begin{tabular}[c]{@{}c@{}}FGSM \\ (8/255)\end{tabular} & \begin{tabular}[c]{@{}c@{}}PGD-8\\ (4/255)\end{tabular} & \begin{tabular}[c]{@{}c@{}}PGD-16\\ (4/255)\end{tabular} & \begin{tabular}[c]{@{}c@{}}CW-100\\ (c=0.01)\end{tabular} & \begin{tabular}[c]{@{}c@{}}CW-100\\ (c=0.05)\end{tabular} & \textbf{Mean $\pm$ S.d.}   \\ \hline
$\sigma$=0.1    & 71.90                                                   & 32.54                                                   & \textbf{23.31}                                           & 79.58                                                     & 71.96                                                     & 55.04 $\pm$ 23.19          \\ \hline
$\sigma$=0.01   & 71.56                                                   & 34.04                                                   & 25.96                                                    & 80.68                                                     & 72.00                                                     & 56.85 $\pm$ 22.31          \\ \hline
$\sigma$=0.005  & 71.31                                                   & 27.79                                                   & 20.64                                                    & 80.46                                                     & 71.58                                                     & 54.36 $\pm$ 24.93          \\ \hline
$\sigma$=0.001  & 70.51                                                   & 27.42                                                   & 17.98                                                    & 79.47                                                     & 67.00                                                     & 52.48 $\pm$ 24.83          \\ \hline
\rowcolor[HTML]{EFEFEF}
$\sigma$=0.05  & \textbf{74.18}                                          & \textbf{32.12}                                          & 22.12                                                    & \textbf{81.39}                                            & \textbf{74.72}                                            & \textbf{56.91 $\pm$ 24.66} \\ \hline
\end{tabular}
}
\end{table}

\begin{table}[]
\centering
\caption{Accuracy (\%) on CIFAR-10 based on WRN-28-10 trained with the proposed method using various of $alpha$ for the Beta distribution to generate mixing coefficient $\lambda$. }
\label{tab:alpha}
\Huge
\resizebox{.48\textwidth}{!}{
\begin{tabular}{c|ccccc|c}
\hline
Alpha          & \begin{tabular}[c]{@{}c@{}}FGSM \\ (8/255)\end{tabular} & \begin{tabular}[c]{@{}c@{}}PGD-8\\ (4/255)\end{tabular} & \begin{tabular}[c]{@{}c@{}}PGD-16\\ (4/255)\end{tabular} & \begin{tabular}[c]{@{}c@{}}CW-100\\ (c=0.01)\end{tabular} & \begin{tabular}[c]{@{}c@{}}CW-100\\ (c=0.05)\end{tabular} & \textbf{Mean $\pm$ S.d.}   \\ \hline
$\alpha$=2.0   & 71.27                                                   & 31.85                                                   & \textbf{22.39}                                           & 79.18                                                     & 70.50                                                     & 55.04 $\pm$ 23.19          \\ \hline
\rowcolor[HTML]{EFEFEF}
$\alpha$=1.0            & \textbf{74.18}                                          & \textbf{32.12}                                          & 22.12                                                    & \textbf{81.39}                                            & \textbf{74.72}                                            & \textbf{56.91 $\pm$ 24.66} \\ \hline
\end{tabular}
}
\vspace{-5mm}
\end{table}

\subsection{Performance on OOD Detection}
To validate the effectiveness of the proposed method for OOD detection, we compare the proposed method with state-of-the-art OOD detection approaches \cite{neal2018open,yoshihashi2019classification,liu2019large,zaeemzadeh2021out} on a public benchmarking dataset. Since MC sampling is a technology that improves performance by sacrificing computational cost (many times higher than other methods), Uni. Sub. \cite{zaeemzadeh2021out} with MC sampling is regarded as an upper bound for OOD detection methods. To ensure the fairness for comparison, we reproduce Uni. Sub. \cite{zaeemzadeh2021out} without employing any MC sampling, denoted as Uni. Sub. w/o MC in Table \ref{tab:OOD}. Using Dynamic Feature Aggregation as a regularization, the CNN model gains an improvement of $\sim3\%$ for F1 score. In particular, when the model trained on CIFAR-10 and tested on LSUN-R, the proposed method can improve the F1 score from $0.907$ to $0.937$, which significantly surpasses the other competing methods. The experimental results further prove the effectiveness of Dynamic Feature Aggregation and give a strong insight for the potential application on OOD detection task.


\begin{table}[tbp]
\caption{Comparison of F1 score for OOD Detection based on WRN-28-10 with various test sets. TIN-C, TIN-R, LSUN-C and LSUN-R refer to the test set of Tiny ImageNet-Crop, Tiny ImageNet-Resize, LSUN-Crop and LSUN-Resize, respectively. $\dagger$ stands for the result reproduced by the open source code.  }
\label{tab:OOD}
\Huge
\resizebox{.48\textwidth}{!}{
\begin{tabular}{ccccc}
\hline
\multicolumn{1}{c|}{In-Distribution Dataset}      & \multicolumn{4}{c}{CIFAR10}                                       \\ \hline
\multicolumn{1}{c|}{OOD Dataset}                  & TIN-C          & TIN-R          & LSUN-C         & LSUN-R         \\ \hline
\multicolumn{1}{c|}{Softmax Pred. (ICLR'2017)\cite{hendrycks2016baseline}}                & 0.803          & 0.807          & 0.794          & 0.815          \\
\multicolumn{1}{c|}{Counterfactual (ECCV'2018)\cite{neal2018open}}               & 0.636          & 0.635          & 0.650          & 0.648          \\
\multicolumn{1}{c|}{CROSR (CVPR'2019)\cite{yoshihashi2019classification}}                        & 0.733          & 0.763          & 0.714          & 0.731          \\
\multicolumn{1}{c|}{OLTR (CVPR'2019)\cite{liu2019large}}                         & 0.860          & 0.852          & 0.877          & 0.877          \\ \hline
\multicolumn{1}{c|}{Uni. Sub. w/o MC $\dagger$ (CVPR'2021)\cite{zaeemzadeh2021out}}        & 0.890          & 0.886          & 0.897          & 0.907          \\ \hline
\rowcolor[HTML]{EFEFEF}
\multicolumn{1}{c|}{\cellcolor[HTML]{EFEFEF}Ours} & \textbf{0.922} & \textbf{0.911} & \textbf{0.934} & \textbf{0.937} \\ \hline
\multicolumn{5}{c}{Methods using MC sampling}                                                                          \\ \hline
\multicolumn{1}{c|}{Uni. Sub. (CVPR'2021)\cite{zaeemzadeh2021out}}        & 0.930          & 0.936          & 0.962          & 0.961          \\ \hline
\multicolumn{5}{c}{Methods which adopt OOD samples for validation and fine-tuning}                                   \\ \hline
\multicolumn{1}{c|}{ODIN (ICLR'2018)\cite{liang2017enhancing}}                         & 0.902          & 0.926          & 0.894          & 0.937          \\
\multicolumn{1}{c|}{Mahalanobis (NIPS'2018) \cite{lee2018simple}}                  & 0.985          & 0.969          & 0.985          & 0.975          \\ \hline
\end{tabular}
}
\vspace{-5mm}
\end{table}

\section{Conclusion and Discussion}
In this paper, we proposed a novel regularization, denoted as Dynamic Feature Aggregation, to improve the robustness against adversarial attacks. By compressing the embedding space of CNN based model, less space is left for the generation of adversarial samples. In particular, the convex combination of two given samples is regarded as their pivot. The model is required to minimize the distance between the given samples and the pivots in the embedding space. With such a constraint, the CNN based model can obtain a compact feature space, but may struggle in avoiding a trivial solution. To alleviate the issue, we then proposed an orthogonal classifier to improve the diversity of the extracted features. Integrating the two components into the CNN based model, the learning objective can be explained as compressing embedding space without sacrificing the capacity of generalization. The effectiveness of the proposed method is validated upon both theoretical analysis and empirical results. Below we briefly discuss some limitations of this study:

\textit{Choice of the noise term.} In order to alleviate over-fitting, we add a noise term to the learning objective, which is introduced in Sec. \ref{sec:DFA}. Based on the discussion in Sec. \ref{sec:adversarial}, the final performance of the proposed method is not sensitive to the factor. However, training or auto-adapting those terms should be a worthwhile potential direction for our future work.

\textit{OOD test.} The proposed method is a regularization, which relies on the specific reference to achieve OOD detection. However, the OOD detection baseline framework adopted in this paper focuses on the class-wise feature, which might not be the best choice for the proposed method. We plan to integrate the proposed method to more state-of-the-art OOD detection methods in the future.


{\small
\bibliographystyle{ieee_fullname}
\bibliography{myreference}
}
\end{document}